\newcommand{\CLASSINPUTinnersidemargin}{18mm}
\newcommand{\CLASSINPUToutersidemargin}{12mm}
\newcommand{\CLASSINPUTtoptextmargin}{20mm}
\newcommand{\CLASSINPUTbottomtextmargin}{25mm}
\documentclass[10pt,conference,twoside,a4paper]{IEEEtran}

\usepackage[utf8]{inputenc}

\usepackage{times}

\usepackage[table,xcdraw]{xcolor}

\usepackage{abstract}
\renewcommand{\abstractname}{}    

\usepackage{graphicx}
\usepackage{subfigure}
\usepackage{svg}
\DeclareGraphicsExtensions{.png,.eps,.ps,.pdf,.svg}

\usepackage{fancyhdr}
\usepackage{kantlipsum}

\usepackage{url}

\usepackage[left=2cm,top=2.5cm,right=2cm,bottom=2.5cm]{geometry}

\usepackage[]{color}

\usepackage{eso-pic}

\begin{document}
\AddToShipoutPictureBG*{%
  \AtPageUpperLeft{%
    \setlength\unitlength{1in}%
 	\hspace{2cm}
 	 	\makebox(0,-2)[l]{
			\begin{tabular}{l r} 
			\multicolumn{1}{p{12cm}}{\vspace{-0.3cm}\includegraphics[scale=0.13]{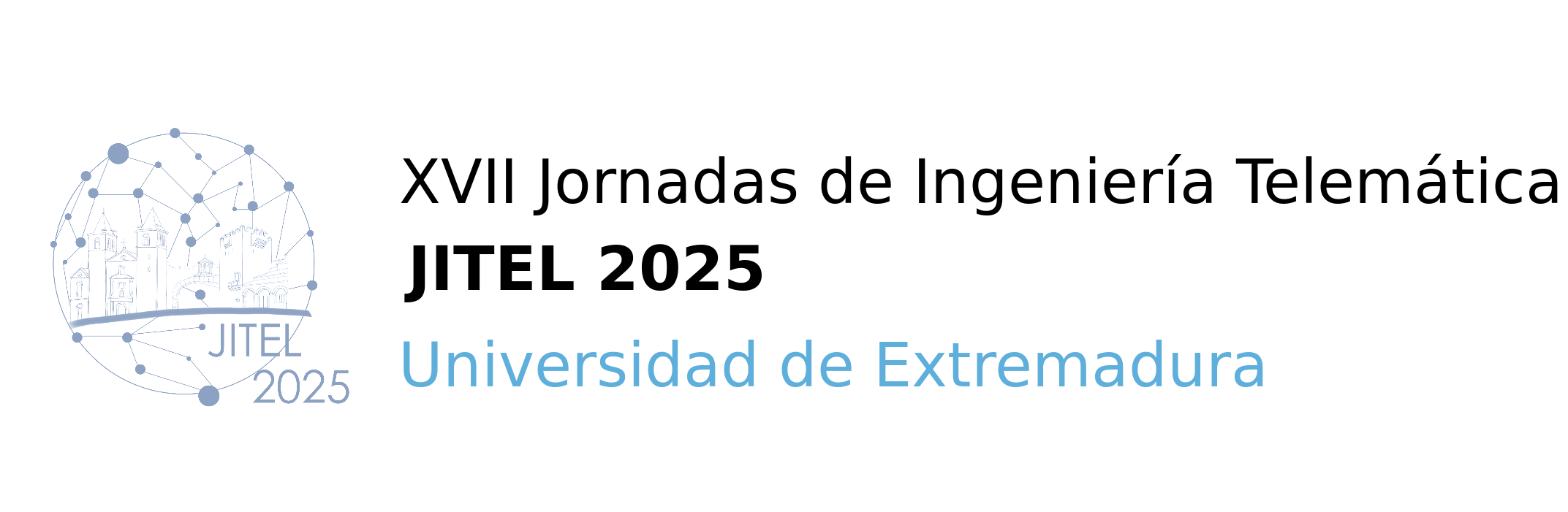}} & \multicolumn{1}{p{4cm}}{\raggedleft\small\usefont{T1}{phv}{m}{it} Actas de las XVII Jornadas de Ingeniería Telemática\\ (JITEL 2025),\\ Cáceres (España), \\12-14 de noviembre de 2025. \vspace{0.2cm} \\ISBN} \tabularnewline 
			\end{tabular}
 }
}}

\AddToShipoutPictureBG*{%
  \AtPageLowerLeft{%
    \setlength\unitlength{1in}%
    \hspace*{\dimexpr0.5\paperwidth\relax}
    \makebox(0,1.3)[c]{\footnotesize\usefont{T1}{phv}{m}{} This work is licensed under a \underline{\textcolor{blue}{Creative Commons 4.0 International License}} (CC BY-NC-ND 4.0)}%

}}

\title{\vspace{2cm}Sistema de Reconocimiento Facial Federado en Conjuntos Abiertos basado en OpenMax}

\author{\IEEEauthorblockA{Ander Galván, Marivi Higuero, Jorge Sasiain, Eduardo Jacob}
\IEEEauthorblockA{
Departamento de Ingeniería de Comunicaciones, Universidad del País Vasco (UPV/EHU),  48013 Bilbao, España \\
\{ander.galvan, marivi.higuero, jorge.sasiain, eduardo.jacob\}@ehu.eus}
}

\maketitle

\begin{abstract}
\textbf{El reconocimiento facial impulsado por Inteligencia Artificial ha demostrado una alta precisión en algunos escenarios y aplicaciones. Sin embargo, presenta desafíos relacionados con la privacidad y la identificación de personas, especialmente considerando que pueden aparecer sujetos desconocidos para el sistema que lo implementa. En este trabajo, se propone el diseño, implementación y evaluación de un sistema de reconocimiento facial en un escenario de aprendizaje federado, orientado a conjuntos abiertos. Concretamente, se diseña una solución basada en el algoritmo OpenMax para escenarios de aprendizaje federado. La propuesta emplea el intercambio de los vectores de activación promedio y distancias locales para identificar de manera eficaz tanto personas conocidas como desconocidas. Los experimentos realizados demuestran la implementación efectiva de la solución propuesta.}
\end{abstract}


\section{\uppercase{Introducción}}

En los últimos años, la Inteligencia Artificial (IA), específicamente el Deep Learning (DL), ha adquirido una importancia significativa en diversos sectores. Entre sus aplicaciones más relevantes, los sistemas de reconocimiento facial impulsados por IA se han consolidado como herramientas claves para la identificación y verificación biométrica de personas en distintos escenarios y aplicaciones. Gracias al uso de Deep Learning, estos sistemas ejecutan tareas que emulan procesos cognitivos humanos, logrando reconocer rostros con alta precisión y adaptándose a condiciones ambientales complejas. Esta capacidad ha favorecido su implementación en ámbitos como el control de acceso, la autenticación biométrica, la videovigilancia y los pagos electrónicos.

Sin embargo, el despliegue de estos sistemas presenta considerables desafíos relacionados con la privacidad de datos cuando se dispone de escenarios de aprendizaje colaborativo. En este contexto, el aprendizaje federado (Federated Learning, FL) surge como una solución efectiva: una estrategia de entrenamiento de modelos que evita la centralización de datos, mantiene la información en los dispositivos locales y realiza agregaciones de conocimiento de forma privada. Este enfoque contribuye a preservar la privacidad de los usuarios y a cumplir con las normativas de protección de datos vigentes \cite{rgpd2016}.

Asimismo, los sistemas de reconocimiento facial deben adaptarse, debido a que no todas las identidades están disponibles durante el proceso de entrenamiento. En muchos escenarios, es fundamental que el sistema no solo identifique con precisión a personas conocidas, sino que también sea capaz de reconocer imágenes pertenecientes a personas desconocidas. Este problema se aborda mediante el enfoque de Reconocimiento en Conjuntos Abiertos (Open-Set Recognition, OSR), que exige que el modelo pueda detectar imágenes que no correspondan a ninguna de las identidades previamente conocidas.

En este contexto, el presente estudio contribuye mediante el diseño, implementación y evaluación de un sistema de reconocimiento facial en un escenario de aprendizaje federado orientado a conjuntos abiertos, garantizando la precisión y privacidad de los datos. Específicamente, se emplea un escenario de aprendizaje federado a lo largo de todo el proceso, en el cual, en primer lugar, se entrena y evalúa un modelo utilizando imágenes de un conjunto de personas conocidas. Posteriormente, se lleva a cabo el diseño de la solución propuesta basada en el algoritmo OpenMax \cite{Bendale2015} con el objetivo de ajustar el modelo para la identificación y clasificación de imágenes de personas no incluidas en el conjunto de entrenamiento previo. Finalmente, se describe la implementación del sistema y los resultados de su evaluación.

El resto de este artículo está estructurado de la siguiente manera. La sección II presenta una visión general de los diferentes ámbitos que componen el trabajo. La sección III detalla el diseño del sistema de reconocimiento facial desarrollado. La sección IV describe la implementación de la solución planteada, mientras que la sección V presenta los resultados obtenidos. Finalmente, la sección VI ofrece un resumen de las principales conclusiones y las líneas de trabajo futuras.

\section{\uppercase{Fundamentos y Contexto Tecnológico}}
\label{contexto}

El Deep Learning emerge como una técnica poderosa y efectiva cuyo funcionamiento se basa en la simulación del cerebro humano, capaz de detectar objetos y personas, reconocer voces, traducir textos, entre otras tareas, y cuya ventaja más significativa es la capacidad para aprender sin supervisión humana \cite{Somayya2022}. Dentro de este enfoque, destacan las redes neuronales convolucionales (Convolutional Neural Network, CNN), un tipo de modelo especializado en el procesamiento de datos estructurados en forma matricial, como las imágenes. Estas redes están generalmente conformadas por capas convolucionales, capas de \textit{pooling} y funciones de activación no lineales, las cuales se encargan de la extracción de características. Posteriormente, pueden incluir capas completamente conectadas y, en caso de tareas de clasificación, se emplea una función de activación SoftMax para interpretar las salidas basándose en las características aprendidas \cite{Alzubaidi2021}.

Por otro lado, el aprendizaje federado permite entrenar modelos de forma colaborativa entre distintas entidades sin la necesidad de compartir datos locales, asegurando así la protección de la privacidad de los datos \cite{Li2023}. Este proceso involucra a los clientes, responsables de los datos, y a un servidor que supervisa la gestión general. En cuanto al proceso de entrenamiento, el servidor inicia el modelo, lo distribuye a los clientes para su entrenamiento local, recibe los parámetros actualizados, los integra y redistribuye el modelo global. Este ciclo se repite de manera iterativa hasta alcanzar la convergencia.

Finalmente, cabe destacar que el entrenamiento de modelos se basa en escenarios de conjunto cerrado, en el cual, en el contexto del reconocimiento facial, todas las imágenes utilizadas en la fase de inferencia corresponden a personas incluidas en el conjunto de entrenamiento, denominadas personas conocidas. Esta condición conlleva que los modelos tiendan a clasificar erróneamente imágenes de personas no presentes en el conjunto de entrenamiento como pertenecientes a personas conocidas, con un alto nivel de confianza \cite{Barcina2024}. En este contexto, OpenMax \cite{Bendale2015} se presenta como una alternativa a la función de activación SoftMax, diseñada para adaptar el comportamiento de los modelos a conjuntos abiertos, permitiendo la clasificación de imágenes de personas con las que no se ha entrenado el modelo, denominadas personas desconocidas, sin asignarlas erróneamente a alguna de las clases conocidas.

Más específicamente, este método implica calcular un vector medio de activación (Mean Activation Vector, MAV) para cada clase conocida, obtenido a partir de las activaciones (las salidas de la última capa totalmente conectada antes de SoftMax) de las muestras correctamente clasificadas durante el entrenamiento. A continuación, se calcula la distancia entre cada muestra correctamente clasificada y su MAV correspondiente, y se ajusta una distribución de Weibull utilizando las $\eta$ mayores distancias (\textit{tail size}). Esta distribución estadística permite estimar la probabilidad de que una nueva muestra no pertenezca a una clase conocida.

Durante la fase de inferencia, el vector de activación de la muestra a clasificar se reajusta en función de estas probabilidades, considerando únicamente las $\alpha$ clases con mayores niveles de activación (\textit{alpha rank}). Además, se calcula una activación para la clase desconocida, la cual se integra al conjunto de clases conocidas. Finalmente, se calculan las probabilidades de las clases conocidas y de la clase desconocida. Si la clase desconocida obtiene la mayor probabilidad, o si la probabilidad de las clases conocidas cae por debajo de un umbral $\epsilon$, la muestra se clasifica como desconocida.

\section{\uppercase{Descripción de la propuesta}}

El presente trabajo aborda el diseño, implementación y evaluación de un sistema de reconocimiento facial orientado a conjuntos abiertos. Dicho trabajo se desarrolla sobre un escenario de aprendizaje federado con el objetivo de garantizar la privacidad de las imágenes faciales de los clientes, sin comprometer la precisión del sistema de reconocimiento facial. 

El sistema consiste en un modelo de reconocimiento facial que se entrena con un conjunto de personas conocidas. El propósito es entrenar un modelo capaz de clasificar dichas personas conocidas. Para analizar dicha capacidad de clasificación, se lleva a cabo una fase de inferencia en conjuntos cerrados, utilizando imágenes no previamente vistas por el modelo de dichas personas conocidas.

Seguidamente, se lleva a cabo una fase de inferencia en conjuntos abiertos, en la que se emplean imágenes tanto de personas conocidas como desconocidas. Además, para realizar una correcta clasificación, en lugar de utilizar la función de activación SoftMax, se implementa la solución propuesta basada en el algoritmo OpenMax \cite{Bendale2015}, la cual permite identificar con mayor precisión a las personas conocidas y etiquetar como desconocidas aquellas que no forman parte del conjunto de entrenamiento.

Sin embargo, tal y como se describe en la sección \ref{contexto}, OpenMax hace uso de las imágenes utilizadas durante el entrenamiento para determinar si una imagen debe clasificarse como perteneciente a una persona conocida o como desconocida. Esto requiere que el servidor tendría que tener acceso a las bases de datos de los clientes, lo cual quebranta los principios del aprendizaje federado, que evita el intercambio de datos sensibles. Alternativamente, cada cliente podría calcular las activaciones correspondientes a las imágenes correctamente clasificadas de cada persona conocida, delegando al servidor la responsabilidad de completar el resto del proceso de OpenMax. Sin embargo, dado que el servidor no tiene por qué ser una entidad confiable, que posee el modelo ya entrenado y que recibe activaciones correspondientes a imágenes de diferentes clientes, la privacidad de las imágenes de estos se puede ver comprometida.

\begin{figure}[h]
    \vspace{-5pt}
    \centerline{\includegraphics[width=6.5cm]{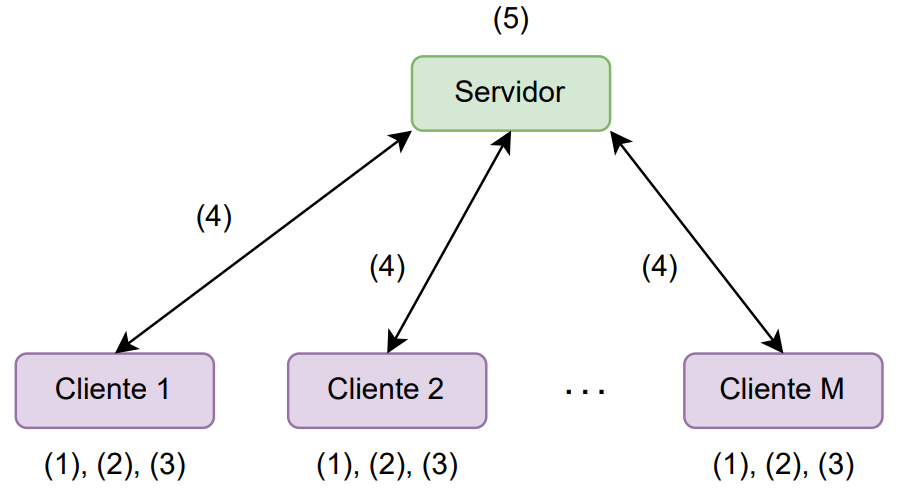}}
    \vspace{-7pt}
    \caption{Solución propuesta basada en OpenMax para escenarios de aprendizaje federado.}
    \label{fig:openmax_fl}
\end{figure}

En consecuencia, la solución planteada hace uso de una propuesta basada en OpenMax para escenarios de aprendizaje federado, como se muestra en la Figura \ref{fig:openmax_fl}. En este caso, una vez entrenado el modelo, cada cliente realiza las siguientes operaciones: (1) calcula las activaciones de las imágenes correctamente clasificadas de cada persona conocida; (2) calcula el MAV correspondiente para cada persona conocida, utilizando las activaciones previamente calculadas; (3) calcula las distancias entre las activaciones de las imágenes correctamente clasificadas y su MAV correspondiente; y (4) envía al servidor tanto el MAV como las distancias calculadas para cada persona conocida. Finalmente, (5) el servidor agrupa la información recibida por persona conocida, calculando la media de los MAV y concatenando las distancias asociadas de los diferentes clientes, con el propósito de completar el proceso de OpenMax de manera centralizada.

En este escenario, el servidor (o las entidades capaces de captar el intercambio de información entre los clientes y el servidor) recibe y almacena únicamente el MAV y las distancias asociadas a cada persona conocida, calculadas de manera local por cada cliente. La transmisión desde cada cliente al servidor del MAV asociado a cada persona conocida, calculado como la media de múltiples activaciones, proporciona un nivel de privacidad superior y disminuye la posibilidad de reconstrucción de características específicas de las imágenes originales, manteniendo así la privacidad de las mismas. Además, las distancias transmitidas no contienen información sensible, ya que son valores derivados de las activaciones. Por consiguiente, compartir dichas distancias no compromete la privacidad de las imágenes.

\section{\uppercase{Implementación}}

En la presente sección se describe la implementación llevada a cabo. Esta misma se organiza en varias subsecciones con el objetivo de ofrecer una descripción clara y completa del escenario.

\subsection{Hardware y software}
La implementación se ha realizado en un servidor Dell PowerEdge R760, equipado con las siguientes especificaciones técnicas: 80 CPU Intel Xeon Silver 4416+, memoria RAM de 128 GB y dos GPU NVIDIA A40 de 48 GB de memoria cada una de ellas. En cuanto al software, el trabajo se ha desarrollado en el lenguaje de programación Python 3.10.12.

\subsection{Datasets} 

Con el propósito de validar y añadir robustez a los resultados obtenidos, se ha replicado el escenario del presente trabajo utilizando dos bases de datos diferentes: MS1M ArcFace \cite{arcface} y VGGFace2 \cite{vggface2}. 

Por un lado, MS1M ArcFace es una base de datos de imágenes faciales derivada de la base de datos original MS-Celeb-1M \cite{ms1m}, la cual ha sufrido procesos de limpieza y preprocesamiento. Esta versión mejorada contiene aproximadamente 5.8 millones de imágenes correspondientes a 85.000 personas. Por otro lado, VGGFace2 se trata de otra base de datos de imágenes faciales publicada en el año 2018, que engloba aproximadamente 3.3 millones de imágenes de 9.000 personas diferentes. 

\subsection{Distribución de datos entre los clientes}

Considerando que el escenario de aprendizaje federado implementado está formado por cinco clientes y un servidor, se ha seleccionado un conjunto de 70 personas conocidas. Cada cliente dispone de 60 imágenes de cada una de estos personas, implementando así una distribución de datos independiente e idénticamente distribuida (Independent and Identically Distributed, IID).

Para validar la capacidad del modelo para clasificar correctamente las 70 personas conocidas (fase de inferencia en conjuntos cerrados), se han empleado otras 50 imágenes de cada persona conocida. Asimismo, en la fase de inferencia en conjuntos abiertos, a las imágenes de las personas conocidas se les ha añadido una única imagen de cada una de otras 8.000 personas desconocidas para el modelo.

\subsection{Algoritmo de agregación}

El algoritmo de agregación del servidor combina los modelos locales de los clientes, integrando su conocimiento y preservando su privacidad. En este contexto, se ha seleccionado el algoritmo FedAvg \cite{fedavg}, debido a su amplio uso, el cual promedia los parámetros de los modelos locales para generar un modelo global.

\subsection{Métricas}

Para evaluar el rendimiento, en las fases de entrenamiento e inferencia en conjuntos cerrados se ha utilizado la precisión (\textit{accuracy}), debido a la distribución equilibrada de imágenes por clase. En la fase de inferencia en conjuntos abiertos, dado el desequilibrio por la inclusión de 8.000 imágenes de personas desconocidas, se ha hecho uso de la métrica F1 macro, que balancea la evaluación considerando todas las clases por igual.

\subsection{Modelo}

Para concluir con esta sección, cabe destacar que se ha utilizado el modelo InceptionResnetV1, previamente pre-entrenado con la base de datos Casia-WebFace \cite{casia-webface}.

\section{\uppercase{Resultados}}

A continuación, se exponen los resultados obtenidos en las tres fases del trabajo: entrenamiento, inferencia en conjuntos cerrados e inferencia en conjuntos abiertos.

\subsection{Entrenamiento del modelo}

La fase de entrenamiento de los clientes empleando ambas bases de datos ha resultado satisfactoria, alcanzando valores de precisión de entrenamiento cercanos al 100\%. Asimismo, cabe destacar la rápida convergencia del modelo para la clasificación de las 70 personas conocidas, lograda en apenas 5 rondas globales de aprendizaje federado. Esto se debe, en primer lugar, a que el modelo se ha iniciado con pesos pre-entrenados, como se ha mencionado anteriormente, y, en segundo lugar, a que cada cliente ha realizado un entrenamiento local durante 3 rondas por cada ronda global de aprendizaje federado.

\subsection{Inferencia en conjuntos cerrados}

Seguidamente, en la fase de inferencia en conjuntos cerrados, se ha evaluado la capacidad del modelo para clasificar a las 70 personas conocidas, utilizando imágenes no vistas por el modelo durante su entrenamiento. El modelo ha demostrado capacidad para clasificar y distinguir dichas personas, alcanzando valores de precisión del 99.6\% y 98.83\% con las bases de datos MS1M ArcFace y VGGFace2, respectivamente.

\subsection{Inferencia en conjuntos abiertos}

Para concluir con los resultados, al implementar la solución propuesta basada en OpenMax y al incluir imágenes de personas desconocidas para el modelo, los valores de F1 macro han experimentado una ligera disminución en comparación con la fase de inferencia en conjuntos abiertos, manteniéndose por encima del 90\%, alcanzando exactamente valores de 93.58\% y 90.8\% para ambas bases de datos. Esto indica que el modelo, generalmente, ha sido capaz de clasificar correctamente a las personas conocidas durante el entrenamiento, así como distinguir a las personas desconocidas.

\section{\uppercase{Conclusiones}}

Este trabajo presenta el diseño, implementación y evaluación de un sistema de reconocimiento facial en un escenario de aprendizaje federado para conjuntos abiertos. Específicamente, se entrena un modelo en un escenario de aprendizaje federado bajo una distribución IID de los datos. Posteriormente, se realiza una fase de inferencia en conjuntos cerrados para evaluar la capacidad del modelo para clasificar dichas personas que han participado en el entrenamiento del mismo. Finalmente, se lleva a cabo una fase de inferencia en conjuntos abiertos para evaluar si el modelo identifica correctamente a personas desconocidas sin perder precisión en la clasificación de las conocidas. Para esta última fase, se ha propuesto una solución basada en el algoritmo OpenMax para escenarios de aprendizaje federado.

Los resultados obtenidos demuestran que el modelo entrenado mediante el escenario de aprendizaje federado desarrollado es capaz de clasificar con precisión a las personas que han participado en dicho proceso, así como de identificar a quienes no lo han hecho, todo ello manteniendo la privacidad de los datos de los clientes. En concreto, en este trabajo se han alcanzado valores de precisión de clasificación en conjuntos abiertos del 93.58\% y 90.8\%. 

Asimismo, las principal contribución de este trabajo es la definición de una solución de un sistema de reconocimiento facial basado en aprendizaje federado y orientado a conjuntos abiertos, empleando OpenMax como base. Cabe destacar que este trabajo resulta especialmente relevante en aplicaciones como sistemas distribuidos de control de acceso biométrico, donde múltiples dispositivos colaboran en el entrenamiento sin compartir datos sensibles. De esta forma, se garantiza la privacidad de los usuarios mientras se mantiene una alta precisión de identificación facial.

Por último, se considera que el trabajo presentado contribuye en la investigación de los ámbitos del aprendizaje federado y el reconocimiento facial en conjuntos abiertos, al integrar de manera conjunta ambos enfoques en el mismo trabajo proporcionando precisión y privacidad. En el futuro, se explorará la utilización de técnicas de protección de la privacidad, tales como la encriptación homomorfica y la privacidad diferencial.

\section*{\uppercase{Agradecimientos}}

Este trabajo ha sido financiado por el proyecto del Ministerio de Ciencia de España a través del proyecto AIBioSurv-Tech (Biosurveillance through Artificial Intelligence (AI) in the post-COVID era: Implications for architecture and cybersecurity), con referencia TED2021-129975B-C22.

\bibliographystyle{IEEEtran}
\bibliography{jitel25}

\end{document}